\documentclass[10pt,twocolumn,letterpaper]{article}

\usepackage{cvpr}              

\usepackage{graphicx}
\usepackage{amsmath}
\usepackage{amssymb}
\usepackage{booktabs}
\usepackage{lipsum} 
\usepackage{graphicx}
\usepackage{booktabs} 
\usepackage{enumitem}
\usepackage{multirow}
\usepackage{multicol}
\usepackage{bbm}
\usepackage[ruled, linesnumbered, noend]{algorithm2e}
\usepackage{algpseudocode}
\usepackage{enumitem}
\usepackage{bm}
\usepackage{cases}

\newcommand{\fout}[0]{f^+}
\newcommand{\fin}[0]{f^-}

\usepackage[pagebackref,breaklinks,colorlinks]{hyperref}

\usepackage[capitalize]{cleveref}
\crefname{section}{Sec.}{Secs.}
\Crefname{section}{Section}{Sections}
\Crefname{table}{Table}{Tables}
\crefname{table}{Tab.}{Tabs.}

\def\cvprPaperID{1159} 
\def\confName{CVPR}
\def\confYear{2023}

\begin{document}

\title{DepGraph: Towards Any Structural Pruning}
\author{\bf Gongfan Fang$^1$ \quad Xinyin Ma$^1$ \quad Mingli Song$^2$ \quad Michael Bi Mi$^3$ \quad Xinchao Wang$^1$\thanks{Corresponding author}\\
{National University of Singapore$^1$ \quad Zhejiang University$^2$ \quad Huawei Technologies Ltd.$^3$} \\
{\tt\small gongfan@u.nus.edu, maxinyin@u.nus.edu,  
xinchao@nus.edu.sg} \\
{\url{https://github.com/VainF/Torch-Pruning}}
}
\maketitle

\begin{abstract}
Structural pruning enables model acceleration by removing structurally-grouped parameters from neural networks. However, the parameter-grouping patterns vary widely across different models, making architecture-specific pruners, which rely on manually-designed grouping schemes, non-generalizable to new architectures. In this work, we study a highly-challenging yet barely-explored task, any structural pruning, to tackle general structural pruning of arbitrary architecture like CNNs, RNNs, GNNs and Transformers. The most prominent obstacle towards this goal lies in the structural coupling, which not only forces different layers to be pruned simultaneously, but also expects all removed parameters to be consistently unimportant, thereby avoiding structural issues and significant performance degradation after pruning. To address this problem, we propose a general and {fully automatic} method, \emph{Dependency Graph} (DepGraph), to explicitly model the dependency between layers and comprehensively group coupled parameters for pruning. In this work, we extensively evaluate our method on several architectures and tasks, including ResNe(X)t, DenseNet, MobileNet and Vision transformer for images, GAT for graph, DGCNN for 3D point cloud, alongside LSTM for language, and demonstrate that, even with a simple norm-based criterion, the proposed method consistently yields gratifying performances.
\end{abstract}

\section{Introduction}

The recent emergence of edge computing applications  calls for
the necessity for deep neural compression~\cite{hinton2015distilling,han2015learning,wu2016quantized,yang2022deep,yang2022KF,Ye2023CVPR,yang2020CVPR,liu2022dataset,LiuSonghua2023CVPR,ruonanDCReviewArxiv22,jing2021meta}. Among the many network compression paradigms, 
pruning has proven itself to be highly effective and practical~\cite{liang2021pruning,lin2020hrank,wang2021accelerate,gao2021network,yu2018nisp,wang2020neural,park2020lookahead,ding2021resrep}.
The goal of network pruning is to 
remove redundant parameters from a given network
to lighten its size and potentially speed up the inference.
Mainstream pruning approaches can be roughly categorized into
two schemes, 
\emph{structurual pruning}~\cite{li2016pruning,ding2019centripetal,you2019gate} and \emph{unstructurual pruning}~\cite{guo2016dynamic,park2020lookahead,dong2017learning}.
The core difference between the two lies in that, 
structural pruning changes the structure of neural 
networks by physically removing grouped parameters, 
while unstructural pruning 
conducts zeroing on partial weights 
without modification to the network structure.
Compared to unstructural ones, structural pruning 
does not rely on specific AI accelerators or software
to reduce memory consumption and computational costs, 
thereby finding a wider domain of applications in practice~\cite{Luo_2020_CVPR,yao2021joint}. 

\begin{figure}
    \centering
    \includegraphics[width=\linewidth]{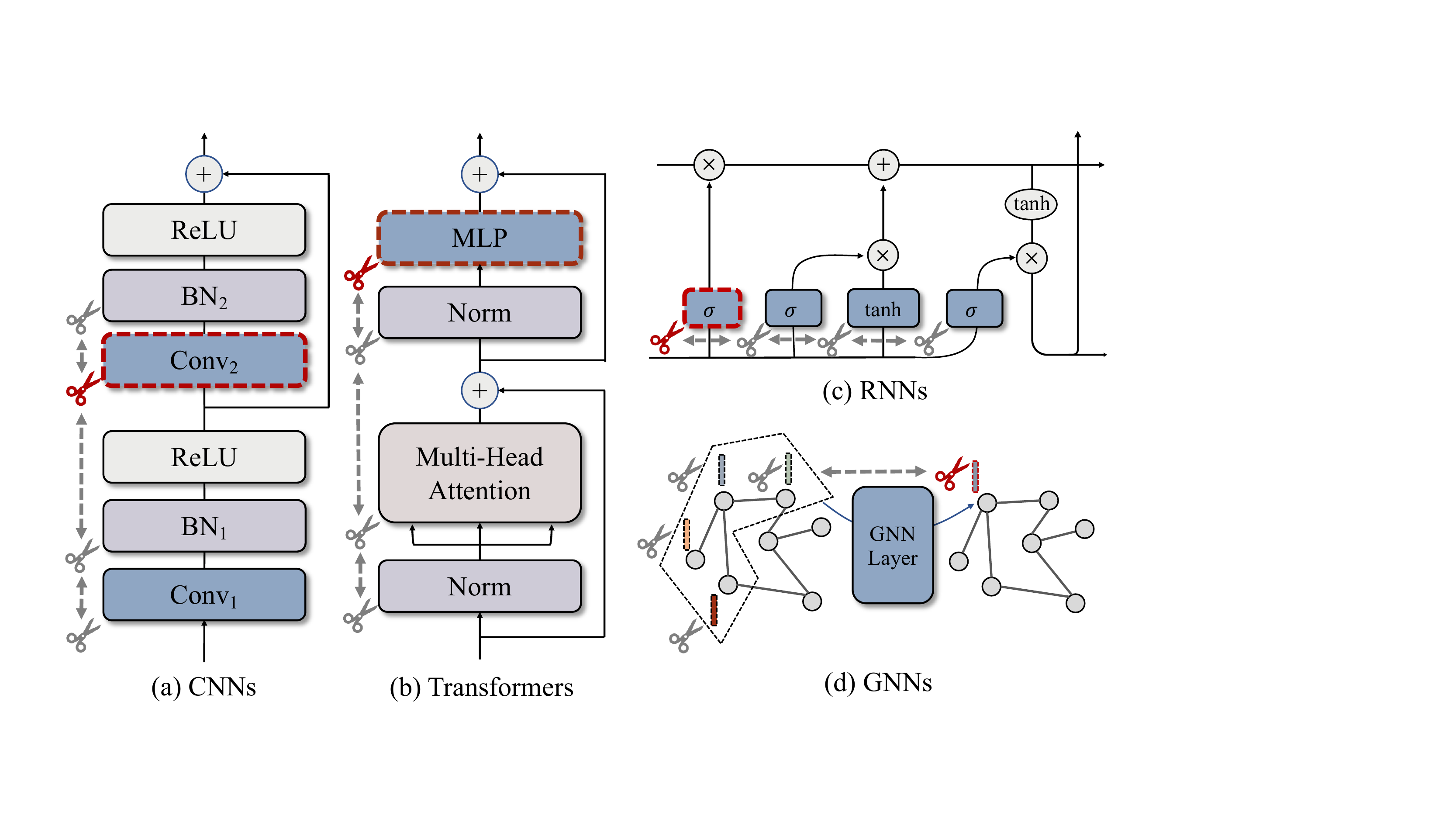}
    \caption{ 
    Parameters from different layers 
    are inherently dependent on each other
    across network architectures, which forces several layers to be pruned simultaneously. For instance, to prune the Conv$_2$ in (a), all other layers \{Conv$_1$, BN$_1$, BN$_2$\} within the block must be pruned as well. In this work, we introduce a generic scheme, termed as Dependency Graph, to explicitly account for such dependencies and execute the pruning of arbitrary architecture in a fully automatic manner.
    }
    \vspace{-2mm}
    \label{fig:intro}
\end{figure}

Nevertheless, 
the nature
of structural pruning \emph{per se}
makes itself a challenging task,
especially for 
modern deep neural networks with coupled and complex internal structures. 
The rationale lies in that, 
deep neural networks are built upon
a large number of basic modules like convolution, 
normalization, or activation,
yet these modules, 
either parameterized or not,
are intrinsically coupled through
the intricate connections~\cite{huang2017densely,he2016deep}. 
As a result, even when we seek to
remove only one channel from a CNN
illustrated in Figure~\ref{fig:intro}(a),
we have to take care of
its inter-dependencies to all layers simultaneously, otherwise we will eventually get a broken network. To be exact, the residual connection requires the output of two convolutional layers to share the same number of channels and thus forces them to be pruned together~\cite{he2019filter,ma2019resnet,you2019gate}. The same goes for structural pruning on other architectures like Transformers, RNNs and GNNs as illustrated in Figs.~\ref{fig:intro}(b-d).

Unfortunately, dependency does not only emerge in residual structures, which can be infinitely complex in modern models~\cite{huang2017densely,radosavovic2020designing}. Existing structural approaches have largely relied on case-by-case analyses to handle dependencies in networks~\cite{li2016pruning,ma2019resnet}. 
Despite the promising results achieved,
such a network-specific pruning approach is
effort-consuming.
Moreover, these methods are 
not directly generalizable,
meaning that the manually-designed 
grouping scheme is not transferable to
other network families 
or even the network architectures in
the same family, which in turn, 
greatly limit their industrial applications.

In this paper, we strive for a generic scheme
towards \emph{any structural pruning}, 
where structural pruning
over arbitrary network architectures
is executed in an automatic fashion, 
At the heart of our approach is to
estimate the \emph{Dependency Graph}~(DepGraph), 
which explicitly models the interdependency 
between paired layers in neural networks. 
Our motivation to introduce DepGraph
for structural pruning stems from
the observation that, structural pruning 
at one layer effectively ``triggers'' 
pruning at adjacent layers, 
which further leads to a chain effect like 
\{BN${_2}$$\leftarrow$Conv${_2}$$\rightarrow$BN${_1}$$\rightarrow$Conv${_1}$\} as shown in Figure~\ref{fig:intro}(a).
As such, to trace the dependencies
across different layers,
we may decompose and model the 
dependency chain as a
recursive process, which naturally
boils down to the problem of finding the maximum connected components in the graph, 
and can be solved in $O(N)$ complexity via graph traversal. 

It is also worth noting that in structural pruning, grouped layers are pruned simultaneously, which expects all removed parameters in the same group to be consistently unimportant. This brings certain difficulties to existing importance criteria designed for a single layer~\cite{he2019filter,li2016pruning,lee2020layer,molchanov2019importance}. To be exact, the parameter importance in a single layer no longer reveals correct importance due to the entanglement with other parameterized layers.
To address this problem, we fully leverage the comprehensive ability of dependency modeling powered by DepGraph to design a ``group-level'' importance criterion, which learns consistent sparsity within groups, so that those zeroized groups can be safely removed without too much performance degradation. 

To validate the effectiveness of DepGraph, we apply the proposed method to several popular architectures including CNNs~\cite{ma2019resnet,huang2017densely}, Transformers~\cite{dosovitskiy2020image}, RNNs~\cite{graves2012long,staudemeyer2019understanding}, and GNNs~\cite{velivckovic2017graph}, where competitive performance is achieved compared to state-of-the-art methods~\cite{you2019gate,wang2020neural,ding2021resrep,liu2021group}. For CNN pruning, our method obtains a $2.57\times$ accelerated ResNet-56 model with 93.64\% accuracy on CIFAR, which is even superior to the unpruned model with 93.53\% accuracy. And on ImageNet-1k, our algorithm achieves more than 2$\times$ speed-up on ResNet-50, with only 0.32\% performance lost. \emph{More importantly}, our method can be readily transferred to various popular networks, including ResNe(X)t~\cite{ma2019resnet,xie2017aggregated}, DenseNet~\cite{huang2017densely}, VGG~\cite{simonyan2014very}, MobileNet~\cite{sandler2018mobilenetv2}, GoogleNet~\cite{szegedy2015going} and Vision Transformer~\cite{dosovitskiy2020image}, and demonstrate gratifying results. Besides, we also conduct further experiments on non-image neural networks, including LSTM~\cite{graves2012long} for text classification, DGCNN~\cite{wang2019dynamic} for 3D point cloud, and GAT~\cite{velivckovic2017graph} for graph data, where our method achieves from 8$\times$ to 16$\times$ acceleration without a significant performance drop.

In sum, our contribution is a generic pruning scheme
towards any structural pruning, 
termed as Dependency Graph (DepGraph), 
which allows for automatic parameter grouping and effectively improves the generalizability of structural pruning over various network architectures, including CNNs, RNNs, GNNs and Vision Transformers.

\newcommand{\boldit}[1]{\textbf{\textit{#1}}}
\newcommand{\bW}[0]{\textit{W}}
\newcommand{\bM}[0]{\textit{M}}
\newcommand{\bP}[0]{\boldit{P}}
\newcommand{\bI}[0]{\textit{I}}
\newcommand{\bD}[0]{\boldit{D}}
\newcommand{\bG}[0]{\boldit{G}}
\newcommand{\argmax}{arg\,max}
\newcommand{\argmin}{arg\,min}
\newcommand{\loss}{\mathcal{L}}
\newcommand{\set}[1]{\{#1\}}
\begin{figure*}[t]
    \centering
    \includegraphics[width=\linewidth]{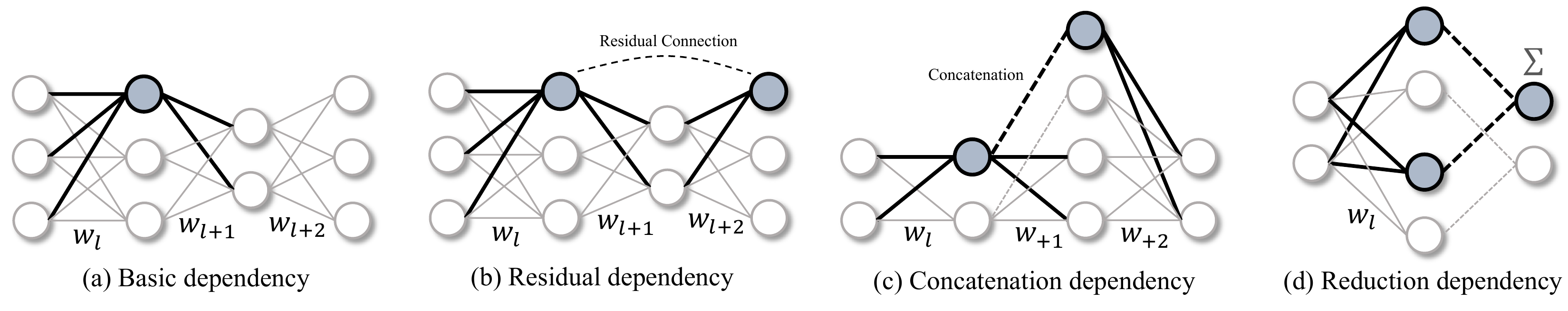}
    \vspace{-2mm}
    \caption{Grouped parameters with inter-dependency in different structures. All highlighted parameters must be pruned simultaneously.}
    \vspace{-2mm}
    \label{fig:parameter_dep}
\end{figure*}

\section{Related Work}

\paragraph{Structural and Unstructural Pruning.}
Pruning has made tremendous progress in the field of network acceleration, as evidenced by various studies in the literature~\cite{liu2017learning,he2017channel,luo2017thinet,li2016pruning,he2019filter,chin2020towards,he2018amc}. Mainstream pruning methods can be broadly categorized into two types: structural pruning~\cite{li2016pruning,ding2019centripetal,you2019gate,liu2021group,you2019gate} and unstructural pruning~\cite{park2020lookahead,dong2017learning,sanh2020movement,lee2019signal}. Structural pruning aims to physically remove a group of parameters, thereby reducing the size of neural networks. In contrast, unstructured pruning involves zeroing out specific weights without altering the network structure.  In practice, unstructural pruning, in particular, is straightforward to implement and inherently adaptable to various networks. however, it often necessitates specialized AI accelerators or software for model acceleration~\cite{han2015deep}. Conversely, structural pruning improves the inference overhead by physically removing parameters from networks, thereby finding a wider domain of applications~\cite{li2016pruning,Luo_2020_CVPR}.
In the literature, The design space of pruning algorithms encompasses a range of aspects, including pruning schemes~\cite{he2017channel,luo2017thinet}, parameter selection~\cite{he2019filter,orseau2020logarithmic,park2020lookahead}, layer sparsity~\cite{lee2020layer,sanh2020movement} and training techniques~\cite{wang2020neural,renda2020comparing}. 
In recent years, numerous robust criteria have been introduced, such as magnitude-based criteria~\cite{he2019filter,lee2020layer, ye2018rethinking} and gradient-based criteria~\cite{liu2021group, lubana2020gradient} and learned sparsity~\cite{ding2021resrep,liu2017learning}. Recently, a comprehensive study has also been conducted to assess the efficacy of various criteria and provide a fair benchmark~\cite{wang2023state}. 

\paragraph{Pruning Grouped Parameters.}  
In complex network structures~\cite{Luo_2020_CVPR,liu2021group,you2019gate,li2016pruning,zhang2021aligned}, dependencies can arise among groups of parameters, necessitating their simultaneous pruning. The pruning of coupled parameters has been a focus of research since the early days of structural pruning~\cite{li2016pruning,liu2017learning,luo2017thinet}. For instance, when pruning two successive convolutional layers, removing a filter from the first layer necessitates the pruning of associated kernels in the subsequent layer~\cite{li2016pruning}. Although it is feasible to analyze parameter dependencies manually, this process can be exceedingly labor-intensive when applied to complicated networks as illustrated in many prior studies~\cite{you2019gate,li2016pruning,zhang2021aligned}. Furthermore, such manual schemes are inherently non-transferable to novel architectures, which severely restricts the applications of pruning. Recently, some pilot works have been proposed to decipher the complex relationships between layers~\cite{liu2021group,you2019gate}. Unfortunately, existing techniques still depend on empirical rules or predefined architectural patterns, rendering them insufficiently versatile for all structural pruning applications. In this study, we present a general approach to resolve this challenge, demonstrating that addressing parameter dependency effectively generalizes structural pruning across a wide array of networks, resulting in satisfactory performance on several tasks.

\newcommand{\dep}[0]{\Leftrightarrow}
\section{Method}

\subsection{Dependency in Neural Networks}\label{sec:dep}

In this work, we focus on structural pruning of \emph{any neural networks} under the restriction of parameter dependency. Without loss of generality, we develop our method upon fully-connected (FC) layers. Let's begin with a linear neural network composed of three consecutive layers as illustrated in Figure \ref{fig:parameter_dep} (a), parameterized by 2-D weight matrices $w_l$, $w_{l+1}$ and $w_{l+2}$ respectively. This simple neural network can be made slim by structural pruning via the removal of neurons. In this case, it is easy to find that some dependencies emerge between parameters, denoted as $w_l \dep w_{l+1}$, which forces $w_l$ and $w_{l+1}$ to be simultaneously pruned. Specifically, to prune the $k$-th neuron that bridges $w_l$ and $w_{l+1}$, both $w_l[k, :]$ and $w_{l+1}[:, k]$ will be removed. In the literature, researchers handle layer dependencies and enable structural pruning on deep neural networks with manually-designed and model-specific schemes~\cite{he2017channel,li2016pruning}. Nevertheless, there are many kinds of dependencies just as illustrated in Figure \ref{fig:parameter_dep} (b-d). It is intractable to manually analyze all dependencies in a case-by-case manner, let alone that simple dependencies can be nested or composed to form more complex patterns. To address the dependency issue in structural pruning, we introduce Dependency Graph in this work, which provides a general and fully-automatic mechanism for dependency modeling.

\subsection{Dependency Graph}

\paragraph{Grouping.}\label{sec:grouping} To enable structural pruning, we first need to group layers according to their inter-dependency. Formally, our goal is to find a grouping matrix $G \in R^{L\times L}$ where $L$ refers to the number of layers in a to-be-pruned network, and $G_{ij}=1$ indicates the presence of dependency between $i$-th layer and $j$-th layer. We let $\text{Diag}(G)=\bm{1}^{1\times L}$ to enable self-dependency for convenience. With the grouping matrix, it is straightforward to find all coupled layers with inter-dependency to the $i$-th layer, denoted as $g(i)$:
\begin{equation}
    g(i) = \{j | G_{ij}=1\}
\end{equation}
Nevertheless, it is non-trivial to estimate the grouping patterns from a neural network due to the fact that modern deep networks may consist of thousands of layers with complicated connections, resulting in a large and intricate grouping matrix $G$. In this matrix, $G_{ij}$ is not only determined by the $i$-th and $j$-th layers but also affected by those intermediate layers between them. Thus, such non-local and implicit relations can not be handled with simple rules in most cases. To overcome this challenge, we do not directly estimate the grouping matrix $G$, and propose an equivalent but easy-to-estimate method for dependency modeling, namely the Dependency Graph, from which $G$ can be efficiently derived.

\begin{figure}
    \centering
    \includegraphics[width=0.95\linewidth]{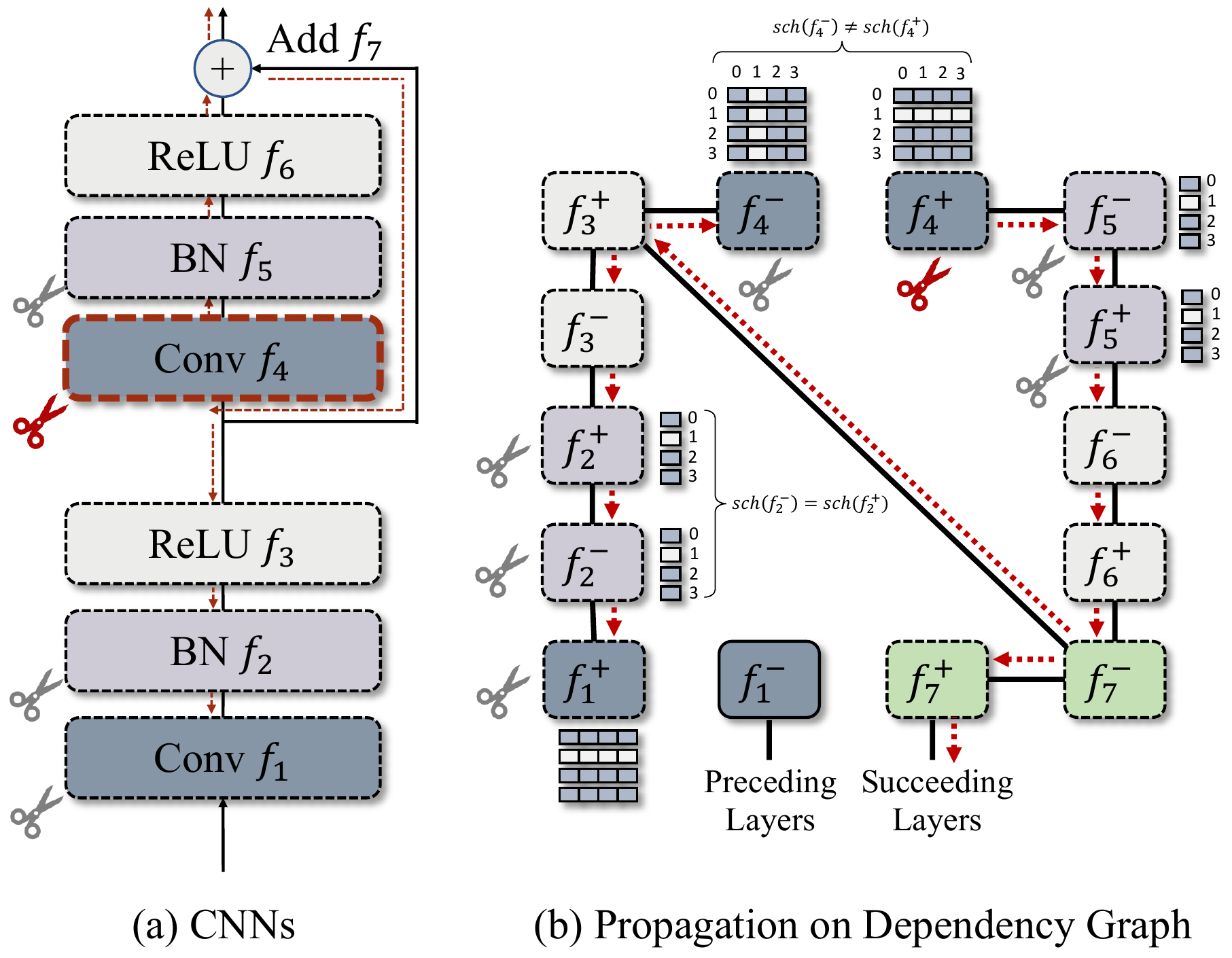}
    \vspace{-2mm}
    \caption{Layer grouping is achieved via a recursive propagation on DepGraph, starting from the $\fout_4$. In this example, there is no Intra-layer Dependency between convolutional input $\fin_4$ and output $\fout_4$ due to the diverged pruning schemes illustrated above.}
    \vspace{-2mm}
    \label{fig:framework}
\end{figure}

\paragraph{Dependency Graph.} Let us begin by considering a group $g=\set{w_1, w_2, w_3}$, which has dependencies $w_1\dep w_2$, $w_2 \dep w_3$, and $w_1\dep w_3$. Upon closer inspection of this dependency modeling, we can observe that there is some redundancy present. For example, the dependency $w_1\dep w_3$ can be derived from $w_1\dep w_2$ and $w_2 \dep w_3$ through a recursive process. Initially, we take $w_1$ as the starting point and examine its dependency on other layers, such as $w_1\dep w_2$. Then, $w_2$ provides a new starting point for recursively expanding the dependency, which in turn "triggers" $w_2 \dep w_3$. This recursive process ultimately ends with a transitive relation, $w_1 \dep w_2 \dep w_3$. In this case, we only need two dependencies to describe the relations in group $g$. Similarly, the grouping matrix discussed in Section \ref{sec:grouping} is also redundant for dependency modeling and thus can be compressed into a more compact form with fewer edges while retaining the same information. We demonstrate that a new graph $D$ that measures the local inter-dependency between adjacent layers, named Dependency Graph, can be an effective reduction for the grouping matrix $G$. The Dependency Graph differs from $G$ in that it only records the dependencies between adjacent layers with direct connections. The Graph $D$ can be viewed as the transitive reduction~\cite{aho1972transitive} of $G$, which contains the same vertices as $G$ but with as few edges as possible. Formally, $D$ is constructed such that, for all $G_{ij}=1$, there exists a path in $D$ between vertex $i$ and $j$. Therefore, $G_{ij}$ can be derived by examing the presence of a path between vertices $i$ and $j$ in $D$.
\newlength\mylen
\newcommand\myinput[1]{%
  \settowidth\mylen{\KwIn{}}%
  \setlength\hangindent{\mylen}%
  \hspace*{\mylen}#1\\}
  
\definecolor{gray}{rgb}{0.7,0.7,0.7}
\algnewcommand{\LineComment}[1]{\State\textcolor{gray}{\(//\) #1}}

\paragraph{Network Decomposition.} 
However, we find that building the dependency graph at the layer level can be problematic in practice, since some basic layers such as fully-connected layers may have two different pruning schemes like $w[k, :]$ and $w[:, k]$ as discussed in Section \ref{sec:dep}, which compress the dimensions of inputs and outputs respectively. Besides, networks also contain non-parameterized operations such as skip connections, which also affect the dependency between layers~\cite{ma2019resnet}. To remedy these issues, we propose a new notation to decompose a network $\mathcal{F}(x; w)$ into finer and more basic components, denoted as $\mathcal{F}=\{f_1, f_2, ..., f_L\}$, where each component $f$ refers to either a parameterized layer such as convolution or a non-parameterized operation such as residual adding. Instead of modeling relationships at the layer level, we concentrate on the dependencies between the inputs and outputs of the layers. Specifically, we denote the input and output of component $f_i$ as $f^-i$ and $f^+i$, respectively. For any network, the final decomposition can be formalized as $\mathcal{F} = \set{\fin_1, \fout_1, ..., \fin_{L}, \fout_{L}}$. This notation facilitates easier dependency modeling and allows different pruning schemes for the same layer.

\paragraph{Dependency Modeling.} Leveraging this notation, we re-sketch the neural network as Equation \ref{eqn:connectivity}, where two principal types of dependencies can be discerned, namely inter-layer dependency and intra-layer dependency, as demonstrated below:
\begin{equation}
\begin{split}
(\fin_1, \underbrace{\fout_1) \leftrightarrow (\fin_2}_{\textit{Inter-layer Dep}}, \fout_2) \dots \leftrightarrow  \underbrace{(\fin_{L}, \fout_{L})}_{\textit{Intra-layer Dep}}
\end{split}
\label{eqn:connectivity}
\end{equation}
The symbol $\leftrightarrow$ signifies the connectivity between two adjacent layers. Examination of these two dependencies yields straightforward but general rules for dependency modeling:
\begin{itemize}[leftmargin=*]
    \item \emph{Inter-layer Dependency}: A dependency $\fin_i \dep \fout_j$ consistently arises in connected layers where $\fin_i \leftrightarrow \fout_j$.
    \item \emph{Intra-layer Dependency}: A dependency $\fin_i \dep \fout_i$ exists if $\fin_i$ and $\fout_i$ share identical pruning schemes, denoted by $sch(\fin_i)=sch(\fout_i)$.
\end{itemize}

\begin{figure*}[t]
    \centering
    \includegraphics[width=0.9\linewidth]{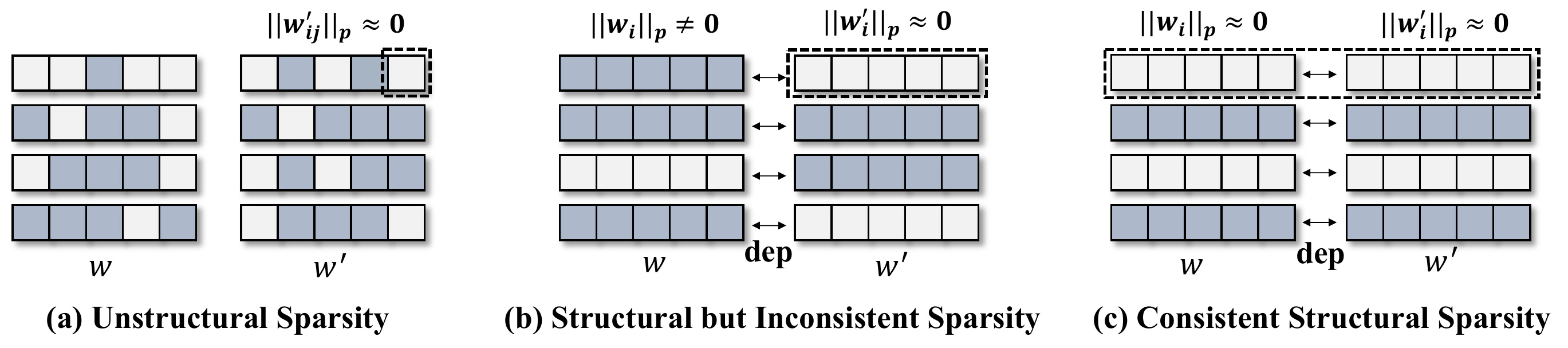}
    \vspace{-2mm}
    \caption{Learning different sparsity schemes to estimate the importance of grouped parameters. Method (a) is used in unstructural pruning which only focuses on the importance of single weight. Method (b) learns structurally sparse layers~\cite{liu2017learning}, but ignores coupled weights in other layers. Our method as shown in (c) learns group sparsity which forces all coupled parameters to zero, so that they can be easily distinguished by a simple magnitude method.}
    \vspace{-2mm}
    \label{fig:sparse_learning}
\end{figure*}

 First, the inter-layer dependency can be readily estimated if the topological structure of the network is known. For connected layers with $\fin_i\leftrightarrow\fout_j$, a dependency consistently exists, because $\fin_i$ and $\fout_j$ correspond to the same intermediate features of the network. The subsequent step involves elucidating the intra-layer dependency. An intra-layer dependency necessitates that both the input and output of a single layer should be pruned simultaneously. Numerous network layers satisfy this condition, such as batch normalization, whose inputs and outputs share the same pruning scheme, denoted as $sch(\fin_i) = sch(\fout_i)$, and thus will be pruned simultaneously as illustrated in Figure~\ref{fig:framework}. In contrast, layers like convolutions have distinct pruning schemes for their inputs and outputs, i.e., $w[:, k, :, :] \neq w[k, :, :, :]$ as shown in Figure \ref{fig:framework}, resulting in $sch(\fin_i) \neq sch(\fout_i)$. In such instances, there is no dependency between the input and output of a convolution layer. Given the aforementioned rules, we can formally establish the dependency modeling as follows:
\begin{equation}\small
D(\fin_i, \fout_j)=\underbrace{\mathbbm{1}\left[\fin_i \leftrightarrow \fout_j\right]}_{\textit{Inter-layer Dep}}   
 \vee \underbrace{\mathbbm{1}\left[i=j 
\land sch(\fin_i)= sch(\fout_j)\right]}_{\textit{Intra-layer Dep}}
\end{equation}
where $\vee$ and $\land$ refer to the logical ``OR'' and ``AND'' operations, and $\mathbbm{1}$ is a indicator function returning ``True'' is the condition holds. The first term examines the Inter-layer Dependency caused by network connectivity, while the second term examines the intra-layer dependency introduced by shared pruning schemes between layer inputs and outputs. It is worth noting that, DepGraph is a symmetric matrix with $D(\fin_i, \fout_j)=D(\fout_j, \fin_i)$. As such, we can examine all input-output pairs to estimate the dependency graph. In Figure~\ref{fig:framework}, we visualize the DepGraph of a CNN block with residual connections. Algorithm \ref{alg:dep_graph} and \ref{alg:grouping} summarize the algorithms for dependency modeling and grouping.

\begin{algorithm}[t]
    \LinesNumberedHidden
   \KwIn{A neural network $\mathcal{F}(x; w)$}
   \KwOut{DepGraph $D(\mathcal{F}, E)$}
   

    $\fin=\{\fin_1, \fin_2, ..., \fin_L\}$ decomposed from the $\mathcal{F}$

    $\fout=\{\fout_1, \fout_2, ..., \fout_L\}$ decomposed from the $\mathcal{F}$
    
    Initialize DepGraph $D = \textbf{0}_{2L\times 2L}$
    
    \For{$i=\set{0, 1, .., L}$}
    {
        \For{$j=\set{0, 1, .., L}$}
        {
            \small 
            $D(\fin_i, \fout_j) = D(\fout_j, \fin_i) =  \underbrace{\mathbbm{1}\left[\fin_i \leftrightarrow \fout_j\right]}_{\textit{Inter-layer Dep}}   \vee \underbrace{\mathbbm{1}\left[i=j \land sch(\fin_i)= sch(\fout_j)\right]}_{\textit{Intra-layer Dep}}$
        }
    }

return $D$
   \caption{Dependency Graph}\label{alg:dep_graph}
\end{algorithm}

\subsection{Group-level Pruning}
In previous sections, we have developed a general methodology for analyzing dependencies within neural networks, which naturally leads to a group-level pruning problem. Assessing the importance of grouped parameters poses a significant challenge to pruning as it involves several coupled layers. In this section, we leverage a simple norm-based criterion~\cite{li2016pruning} to establish a practical method for group-level pruning. Given a parameter group $g=\{w_1, w_2, ..., w_{|g|}\}$, existing criteria like L$_{2}$-norm importance $I(w) = \|w\|_2$ can produce independent scores for each $w\in g$. A natural way to estimate the group importance would be computing an aggregated score $I(g) = \sum_{w\in g}{I(w)}$. Unfortunately, importance scores independently estimated on different layers are likely to be non-additive and thus meaningless due to the divergence of distributions and magnitudes. To make this simple aggregation works for importance estimation, we propose a sparse training method to sparsify parameters at the group level as illustrated in Figure \ref{fig:sparse_learning} (c), so that those zeroized groups can be safely removed from the network. Specifically, for each parameters $w$ with $K$ prunable dimensions indexed by $w[k]$, we introduce a simple regularization term for sparse training, defined as:
\begin{algorithm}[t]
    \LinesNumberedHidden
   \KwIn{DepGraph $D(\mathcal{F}, E)$}
   \KwOut{Groups G}

    $G = \set{}$
    
    \For{ $i = \set{1, 2, ..., 2*\|\mathcal{F}\|}$ }
    {   
        $g = \set{i}$ 
        
       \Repeat{$g^{\prime}=\emptyset$}
       {    
            UNSEEN = $\set{1, 2, ..., 2*\|\mathcal{F}\|} - g$
            
            $g^{\prime} = \{j \in \text{UNSEEN} | \exists k \in g, D_{kj}=1\}$

            $g = g \cup g^{\prime}$
       }
       $G = G \cup \set{g}$
    }
    return $G$
    
   \caption{Grouping}\label{alg:grouping}
\end{algorithm}
\begin{equation}
    \mathcal{R}(g, k) = \sum_{k=1}^K \gamma_k \cdot I_{g,k} = \sum_{k=1}^K \sum_{w\in g}  \gamma_k \|w[k]\|_2^2 
\end{equation}
where $I_{g,k}=\sum_{w\in g} \|w[k]\|_2^2$ represents the importance of the $k$-th prunable dimensions, and $\gamma_k$ refers to the shrinkage strength applied to those parameters. We use a controllable exponential strategy to determine the $\gamma_k$ as follows: 
\begin{equation}
    \gamma_k = 2^{\alpha (I_g^{\text{max}} - I_{g,k}) / (I_g^{\text{max}} - I_g^{\text{min}})}
\end{equation}
where a normalized score is used to control the shrinkage strength $\alpha_k$, varying within the range of $\left[2^0, 2^\alpha\right]$. In this work, we use a constant hyper-parameter $\alpha=4$ for all experiments.  
After sparse training, we further use a simple relative score $\hat{I}_{g,k} = N \cdot I_{g,k} / \sum\set{\text{TopN}(I_g)}$ to identify and remove unimportant parameters. In the experiments section, we show that such a simple pruning method, when combined with consistent sparse training, can achieve comparable performance to modern approaches.

\section{Experiments}

\subsection{Settings} 
This paper focuses on classification tasks and conducts extensive experiments on a variety of datasets, such as CIFAR~\cite{krizhevsky2009learning} and ImageNet~\cite{deng2009imagenet} for image classification, PPI~\cite{hamilton2017inductive} for graph classification, ModelNet~\cite{wu20153d} for 3D classification, and AGNews~\cite{zhang2015character} for text classification. For each dataset, we evaluated our method on several popular architectures, including ResNe(X)t~\cite{ma2019resnet,xie2017aggregated}, VGG~\cite{simonyan2014very}, DenseNet~\cite{huang2017densely}, MobileNet~\cite{sandler2018mobilenetv2}, GoogleNet~\cite{szegedy2015going}, Vision Transformers~\cite{dosovitskiy2020image}, LSTM~\cite{graves2012long}, DGCNNs~\cite{wang2019dynamic}, and Graph Attention Networks~\cite{velivckovic2017graph}. To conduct ImageNet experiments, we use off-the-shelf models from Torchvision~\cite{marcel2010torchvision} as the original models. After pruning, All models will be fine-tuned following a similar protocol as the pre-training stage, with a smaller learning rate and fewer iterations. 

\subsection{Results on CIFAR}

\begin{table}[t]
  \centering
  \resizebox{\linewidth}{!}{
  \small
  \begin{tabular}{c l | c c | c c}
      \toprule
      \bf Model / Data & \bf Method & \bf Base  & \bf Pruned & \bf $\Delta$ Acc. &  \bf Speed Up\\
      \midrule  
      \multirow{17}{*}{\shortstack{ResNet56\\CIFAR10}} 
      & NISP~\cite{yu2018nisp}                  & - & - & -0.03 & 1.76$\times$ \\
      & Geometric~\cite{he2019filter}           & 93.59 & 93.26 & -0.33 & 1.70$\times$ \\
      & Polar~\cite{zhuang2020neuron}    & 93.80 & 93.83 & +0.03 & 1.88$\times$\\
      & CP~\cite{li2016pruning}                  & 92.80 & 91.80 & -1.00 & 2.00$\times$   \\
      & AMC~\cite{he2018amc}                    & 92.80 & 91.90 & -0.90 & 2.00$\times$  \\
      & HRank~\cite{lin2020hrank}              & 93.26 & 92.17 & -0.09 & 2.00$\times$ \\
      & SFP~\cite{he2018soft}                   & 93.59 & 93.36 & -0.23 & 2.11$\times$  \\
      & ResRep~\cite{ding2021resrep} & 93.71 & 93.71 & +0.00 & \bf 2.12$\times$ \\
      & Ours w/o SL                              & 93.53 & 93.46 & -0.07 & 2.11$\times$\\
      & \bf Ours                                & 93.53 & \bf 93.77 & \bf +0.24 & 2.11$\times$ \\
      
      \cmidrule{2-6}
      & GBN (\cite{you2019gate})            & 93.10 & 92.77 & -0.33 & 2.51$\times$ \\
      & AFP (\cite{ding2018auto})           & 93.93 & 92.94 & -0.99 & 2.56$\times$ \\
      & C-SGD (\cite{ding2019centripetal})  & 93.39 & 93.44 & +0.05 & 2.55$\times$ \\
      & GReg-1 (\cite{wang2020neural})      & 93.36 & 93.18 & -0.18 & 2.55$\times$  \\
      & GReg-2 (\cite{wang2020neural})      & 93.36 & 93.36 & -0.00 & 2.55$\times$  \\
      & Ours w/o SL                         & 93.53 &  93.36 & -0.17 &  2.51$\times$ \\
      & \bf Ours & 93.53 & \bf 93.64 & \bf +0.11 & \bf 2.57$\times$ \\
      
      \midrule
      \multirow{7}{*}{\shortstack{VGG19\\CIFAR100}} 
      & OBD (\cite{wang2019eigendamage})    & 73.34 & 60.70 & -12.64 & 5.73$\times$   \\
      & OBD (\cite{wang2019eigendamage})    & 73.34 & 60.66 & -12.68 & 6.09$\times$  \\
      & EigenD (\cite{wang2019eigendamage}) & 73.34 & 65.18 & -8.16 & 8.80$\times$  \\
      & GReg-1 (\cite{wang2020neural})      & 74.02 & 67.55 & -6.67 & 8.84$\times$ \\
      & GReg-2 (\cite{wang2020neural})      & 74.02 & 67.75 & -6.27 & 8.84$\times$ \\
      & Ours w/o SL &  73.50 &  67.60 & -5.44  & 8.87$\times$  \\
      & \bf Ours & 73.50 & \bf 70.39 & \bf -3.11  & 
      \bf 8.92$\times$  \\
      \bottomrule
  \end{tabular} 
  }
  \vspace{-2mm}
  \caption{Pruning results on CIFAR-10 and CIFAR-100.}
  \vspace{-2mm}
  \label{tbl:pruning_cifar}
\end{table}

\paragraph{Performance.}  CIFAR~\cite{krizhevsky2009learning} is a tiny image dataset, which is widely used to verify the effectiveness of pruning algorithms. We follow existing works~\cite{wang2020neural,ding2021resrep} to prune a ResNet-56 on CIFAR-10, and a VGG network on CIFAR-100. As shown in Table \ref{tbl:pruning_cifar}. We report the accuracy of the pruned models as well as their theoretical speedup ratios, defined as $\text{Speed Up}=\frac{\text{MACs}(\text{base})}{\text{MACs}(\text{pruned})}$. Note that baselines like ResRep~\cite{ding2021resrep}, GReg~\cite{wang2020neural} also deploy sparse training for pruning. A key difference between our algorithm and existing sparsity-based algorithms is that our pruner consistently promotes sparsity across all grouped layers, convering convolutions, batch normalizations and fully-connected layers. With this improvement, we are able to take full advantage of the group structures to learn better sparsity, and thus improve the accuracy of pruned models.

\begin{table*}[t]
  \centering
  \small
  \begin{tabular}{c l c c c c c c c c c}
      \toprule
      \multirow{2}{*}{\bf Architecture} & 
      \multirow{2}{*}{\bf Strategy}  &
      \multicolumn{5}{c}{\bf Pruned Accuracy  with Uniform / Learned Sparsity} \\
     \cmidrule{3-7}
      & & \bf 1.5$\times$ & \bf 3.0$\times$ & \bf 6.0$\times$ & \bf 12$\times$ & \bf Avg.  \\
    \midrule 
    \multirow{4}{*}{\shortstack{ResNet-56\\(72.58)}}
    & Random         & 71.49 / 72.07 & 68.52 / 68.16 & 60.35 / 60.25 & 53.21 / 48.01 & 63.39 / 62.15 \\      
    & No grouping    & {71.96} / 72.07 & 67.85 / 67.89 & 62.64 / 63.18 & 54.52 / 53.65  & 64.24 / 64.20 \\  
    & Conv-only             & 71.64 / 71.94 & 68.30 / 69.07 & 62.44 / 62.63 & 53.89 / 54.94 & 64.07 / 64.65 \\
    & \bf Full Grouping  & 71.68 / {\bf 72.57} &  68.70 / {\bf 70.38} &  63.72 / {\bf 65.33} & 55.23 / {\bf  55.92} & 64.83 / {\bf 66.09} \\
    \midrule
    \multirow{4}{*}{\shortstack{VGG-19\\(73.50)}}  
    & Random     & 72.63 / 72.77 & 71.27 / 70.83 & 68.97 / 69.16 & 62.45 / 63.42 & 63.83 / 69.05 \\ 
    & No Grouping             & {73.83} / 55.13 & 71.40 / 53.21 & 69.19 / 50.10 & 65.12 / \, {}$^\dagger$3.87 & 69.14 / 40.58 \\
    & Conv-Only              & 73.32 / 73.22 & 71.38 / 71.80 & 69.66 / 69.85 & 64.69 / 65.95 & 69.76 / 70.21 \\
    & \bf Full Grouping  & 73.11 / {\bf 74.00} &  71.57 / \bf 72.46 &  69.72 / \bf 70.38 &  65.74 / \bf 66.20 &  70.03 / \bf 70.58 \\
    \midrule
    \multirow{4}{*}{\shortstack{DenseNet-121\\(78.73)}}  
    & Random     & 79.04 / 79.43 & 77.86 / 78.62 & 75.47 / 74.52 & 69.26 / 69.64 & 75.41 / 75.80 \\
    &  No Grouping           & 79.31 / 78.91 & 78.08 / 78.62 & {\bf  78.62} / 68.57 & 72.93 / 57.17 & 77.24 / 70.82 \\ 
    & Conv-Only       & 79.18 / {\bf 79.74} & {77.98} / 78.85 & 76.61 / 77.22 & 73.30 / 73.95 & 76.77 / 77.44 \\
    & \bf Full Grouping  & {79.34} / 79.74 & 77.97 / {\bf 79.19} & 77.08 / {77.78} & {\bf 74.77} / 75.29 &  77.29 / {\bf 77.77} \\
    \midrule
    \multirow{4}{*}{\shortstack{MobileNetv2\\(70.80)}} 
    & Random     & 70.90 / 70.69 & 67.75 / 67.54 & 61.32 / 62.26 & 53.41 / 53.97 & 63.35 / 63.62 \\
    & No Grouping             & 71.16 / 71.28 & 69.93 / 68.59 & {66.76} / 37.38  & 60.28 / 28.24 & 67.03 / 51.37  \\
    & Conv-Only              & {71.22} / 71.51 & {70.33} / 70.15 & 66.16 / 66.49 & {61.35} / 63.24 & {67.27} / 67.85 \\
    & \bf Full Grouping  & 71.11 / {\bf 71.67} & 70.06 / {\bf 70.81} & 66.48 / {\bf 68.02} & 60.32 / {\bf 63.37} & 66.99 / {\bf 68.67} \\
    \midrule
    \multirow{4}{*}{\shortstack{GoogleNet\\(77.56)}} 
    & Random     & 77.52 / 77.72 & 76.47 / 76.15 & 74.92 / 74.19 & 69.37 / 69.69 & 74.57 / 74.44 \\
    & No Grouping            & 77.44 / 77.23 & 76.84 / 74.95 & 75.60 / 63.78 & 71.92 / 63.72 &  75.45 / 69.92 \\  
    & Conv-Only              & 77.33 / {\bf 77.62} & 76.68 / 76.92 & 75.66 / 74.98 & 71.90 / 71.87 & 75.49 / 75.35 \\
    & \bf Full Grouping   & 77.91 / 77.76 & 76.90 / {\bf 77.00}& 75.42 / {\bf 75.44} & 71.98 / {\bf 72.88} & 75.53 / {\bf 75.57}  \\
     \bottomrule
  \end{tabular} 
  \caption{Ablation study on CIFAR-100 for different grouping strategies and sparsity configurations. The proposed strategy, full grouping, takes all parameterized layers into account during sparse training, while other strategies only leverage partial layers. Accuracy (\%) of pruned models with uniform layer sparsity or learned layer sparsity is reported. $\dagger$:In some cases, our method over-prunes some dimension to 1, which severely damages the final accuracy.}
  \vspace{-2mm}
  \label{tbl:grouping_strategy}
\end{table*} 

\begin{figure}[t]
    \centering
    \includegraphics[width=\linewidth]{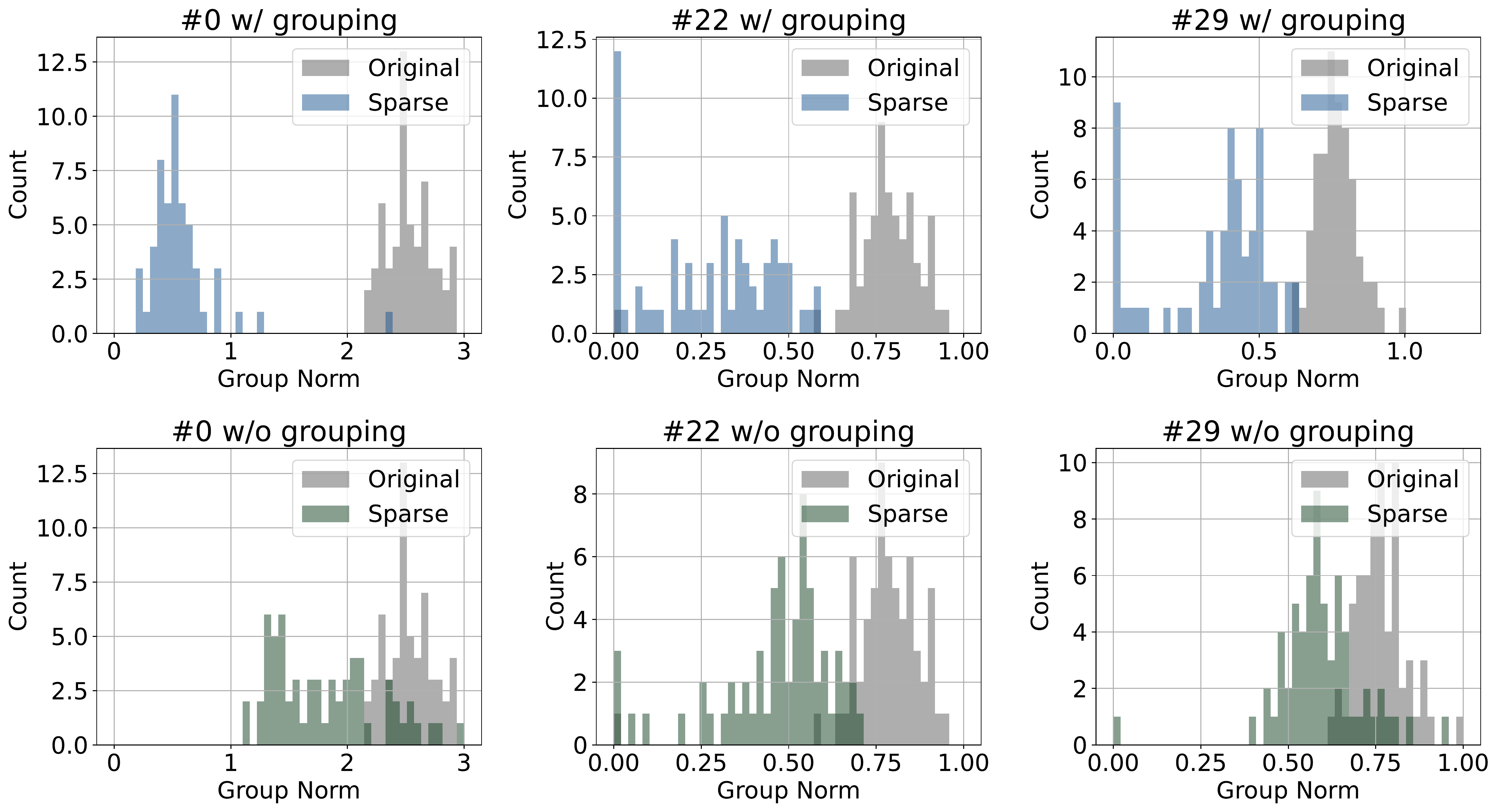}
    \vspace{-2mm}
    \caption{Histogram of group-level sparsity obtained by sparse learning w/ and w/o grouping, which respectively correspond to the strategy (c) and (b) in Figure.\ref{fig:sparse_learning}.}
    \label{fig:cifar_sparsity}
    \vspace{-2mm}
\end{figure}

\paragraph{Distribution of Group Sparsity.} As previously stated, consistent sparsity are important for structural pruning, as it forces all pruned parameters to be consistently unimportant. In Figure \ref{fig:cifar_sparsity}, we visualize the norm of grouped parameters learned by the consistent and inconsistent strategies in Figure~\ref{fig:sparse_learning} (c) and (b). It is easy to find that our method produces strong sparsity at the group level, which would be beneficial for identifying unimportant parameters. However, the inconsistent method that works on different layers independently fails to produce consistent importance across layers, which could result in non-sparse norm at the group level.

\subsection{Ablation Study} 

\paragraph{Grouping Strategy.} To further verify the effectiveness of grouping, we evaluate different strategies on several convolutional networks. Strategies mainly include: 1) No grouping: sparse learning and importance evaluation are performed independently on a single convolutional layer; 2) Conv-only Grouping: all convolutional layers within the group are sparsified in a consistent manner. 3) Full Grouping: All parameterized layers within a group, such as convolutions, batch normalizations, and fully-connected layers, are sparsified consistently. As shown in Table \ref{tbl:grouping_strategy},  When we ignore the grouping information in neural networks and sparsify each layer in isolation, the performance of our approach will degrade significantly, and in some cases even collapse due to over-pruning. The results of Conv-only setting show that grouping partial parameters is beneficial to the final performance, but some useful information in the group is still ignored. Therefore, it is feasible to further improve the pruning accuracy with the full grouping strategy.

\paragraph{Learned Sparsity.} Layer sparsity is also an important factor for pruning, which determines the final structure of pruned neural networks. Table \ref{tbl:grouping_strategy} provides some results about layer sparsity. This work primarily focuses on two types of sparsity, namely uniform sparsity and learned sparsity. With uniform sparsity, the same pruning ratio will be applied to different layers, assuming that redundancy is distributed uniformly through the network. However, previous experiments in Figure \ref{fig:cifar_sparsity} have shown that different layers are not equally prunable. In most cases, the learned sparsity outperforms the uniform one, although sometimes it may over-prune some layers, leading to degraded accuracy.

\paragraph{Generalizability of DepGraph.} Results in Table \ref{tbl:grouping_strategy} also demonstrate the generalizability of our framework, which is able to handle various convolutional neural networks. Moreover, we emphasize that our method is compatible with DenseNet and GoogleNet, which contains dense connections and parallel structures. In the following sections, we will further demonstrate the capability of our framework to more architectures.

\begin{table}[t]
  \centering
  \small
  \resizebox{\linewidth}{!}{
  \begin{tabular}{c l | c c | c c}
      \toprule
      \bf Arch. & \bf Method   & \bf Base & \bf Pruned & \bf $\Delta$ Acc. & \bf MACs  \\
    \hline  
      \bf \multirow{14}{*}{\rotatebox[origin=c]{90}{ResNet-50}} 
      & ResNet-50 & 76.15 & - & - & 4.13 \\
      & ThiNet~\cite{luo2017thinet} & 72.88  & 72.04 & -0.84 & 2.44 \\
      & SSS~\cite{huang2018data}  & 76.12 & 74.18 & -1.94 & 2.82\\ %
      
      & SFP~\cite{he2018soft}   & 76.15 & 74.61 & -1.54 & 2.40 \\ 
      & AutoSlim~\cite{yu2019autoslim}  & 76.10 & 75.60 & -0.50 & 2.00 \\ 
      & FPGM~\cite{he2019filter}   & 76.15 & 75.50 & -0.65 & 2.38 \\ 
      & Taylor~\cite{molchanov2019importance}  & 76.18 & 74.50 & -1.68 & 2.25 \\
      & Slimable~\cite{yu2019slimmable}  &  76.10 & 74.90 & -1.20 &  2.30 \\ 
      & CCP~\cite{peng2019collaborative}  & 76.15 & 75.50 & -0.65 & 2.11 \\
      & AOFP-C1~\cite{ding2019approximated}  & 75.34 & 75.63 & +0.29 &  2.58 \\ 
      & TAS~\cite{dong2019network}  & 77.46 & 76.20 & -1.26 & 2.31 \\ 
      & GFP~\cite{liu2021group}  & 76.79 & 76.42 & -0.37 & 2.04 \\ 
      & GReg-2~\cite{wang2020neural} & 76.13 & 75.36 & -0.77 & 2.77 \\ 
      & Ours & 76.15 & 75.83 & -0.32 & 1.99\\
      \midrule
      \bf \multirow{5}{*}{\rotatebox[origin=c]{90}{DenseNet-121}} &
      DenseNet-121 & 74.44 & - & - & 2.86 \\
      & PSP-1.38G~\cite{schindler2020parameterized}  & 74.35 & 74.05 & -0.30  & 1.38 \\ 
      & PSP-0.58G~\cite{schindler2020parameterized}  & 74.35 & 70.34 & -4.01  & 0.58 \\ 
      & Ours-1.38G & 74.44 & 73.98 & -0.46 & 1.37  \\
      & Ours-0.58G  & 74.44 & 70.13 & -4.31 & 0.57  \\
      \midrule
      \bf \multirow{5}{*}{\rotatebox[origin=c]{90}{Mob-v2}} 
      & Mob-v2 & 71.87 & - & - & 0.33\\
      & NetAdapt~\cite{yang2018netadapt} & - & 70.00 & - & 0.24 \\
      & Meta~\cite{liu2019metapruning}  & 74.70 & 68.20 & -6.50 & 0.14  \\
      & GFP~\cite{liu2021group} & 75.74 & 69.16 & -6.58 & 0.15 \\
      & Ours & 71.87 & 68.46 & -3.41 & 0.15 \\
      \midrule
      \bf \multirow{4}{*}{\rotatebox[origin=c]{90}{NeXt-50}} 
      & ResNeXt-50 & 77.62 & - & - & 4.27 \\
      & SSS~\cite{huang2018data} & 77.57 & 74.98 & -2.59 & 2.43 \\ 
      & GFP~\cite{liu2021group} & 77.97 & 77.53 & -0.44 & 2.11 \\
      & Ours & 77.62 & 76.48 & -1.14 & 2.09  \\
      \midrule
      \bf \multirow{4}{*}{\rotatebox[origin=c]{90}{ViT-B/16}} 
      & VIT-B/16 & 81.07 & - & - & 17.6 \\
      & CP-ViT~\cite{song2022cp}  & 77.91 & 77.36 & -0.55 & 11.7 \\ 
      & Ours+EMA & 81.07 & 79.58 & -1.39 & 10.4 \\
      & Ours & 81.07 & 79.17 & -1.90 & 10.4 \\
      \bottomrule
      
  \end{tabular} 
  }
  \caption{Pruning results on ImageNet.}
  \label{tbl:imagenet_pruning}
\end{table} 
\subsection{Towards Any Structural Pruning}

\begin{figure}
     \centering
     \begin{subfigure}[b]{0.15\textwidth}
         \centering
         \caption{DenseNet-121}
         \includegraphics[width=\textwidth]{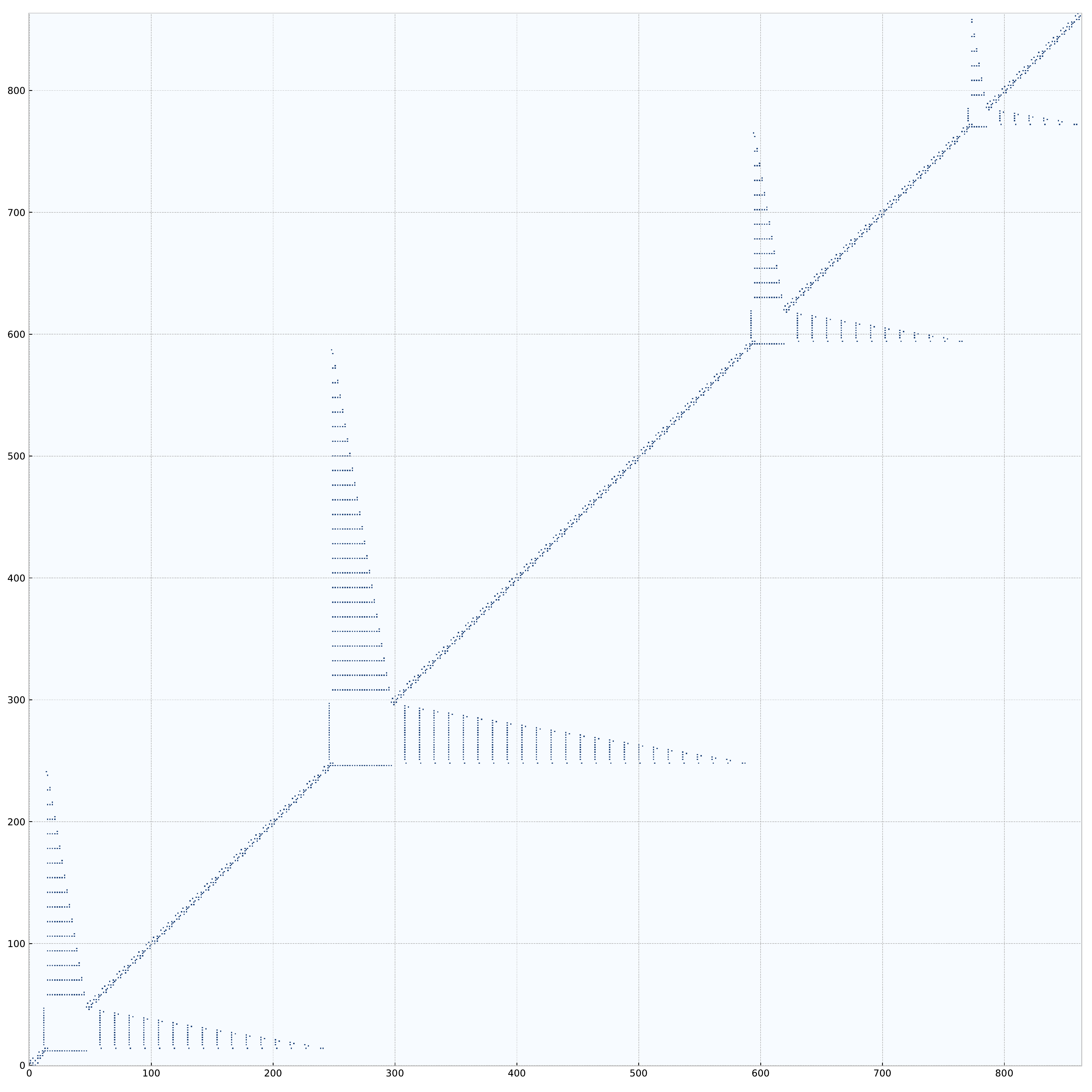}
     \end{subfigure}
     \begin{subfigure}[b]{0.15\textwidth}
         \centering
         \caption{ResNet-18}
         \includegraphics[width=\textwidth]{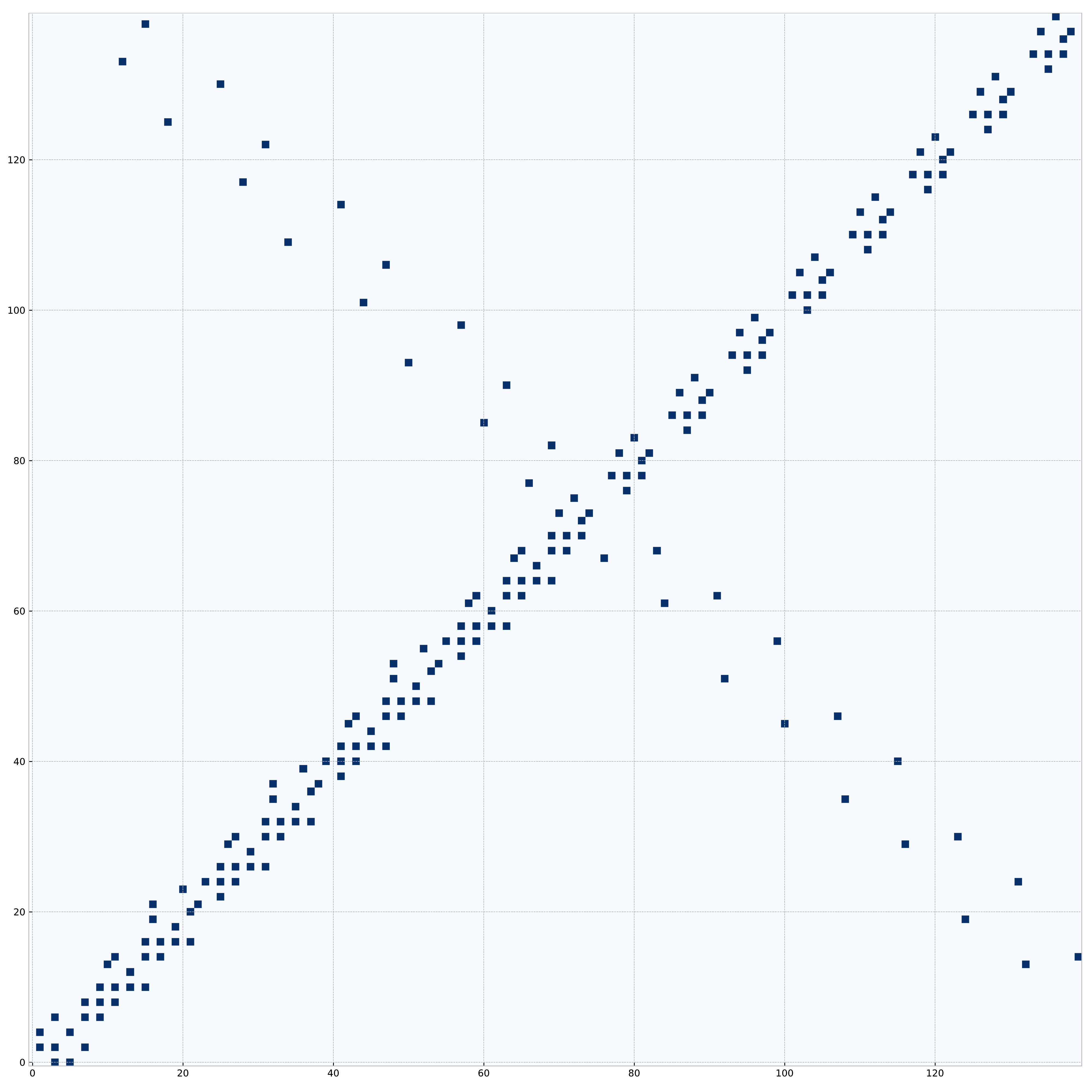}
     \end{subfigure}
     \begin{subfigure}[b]{0.15\textwidth}
         \centering
         \caption{ViT-Base}
         \includegraphics[width=\textwidth]{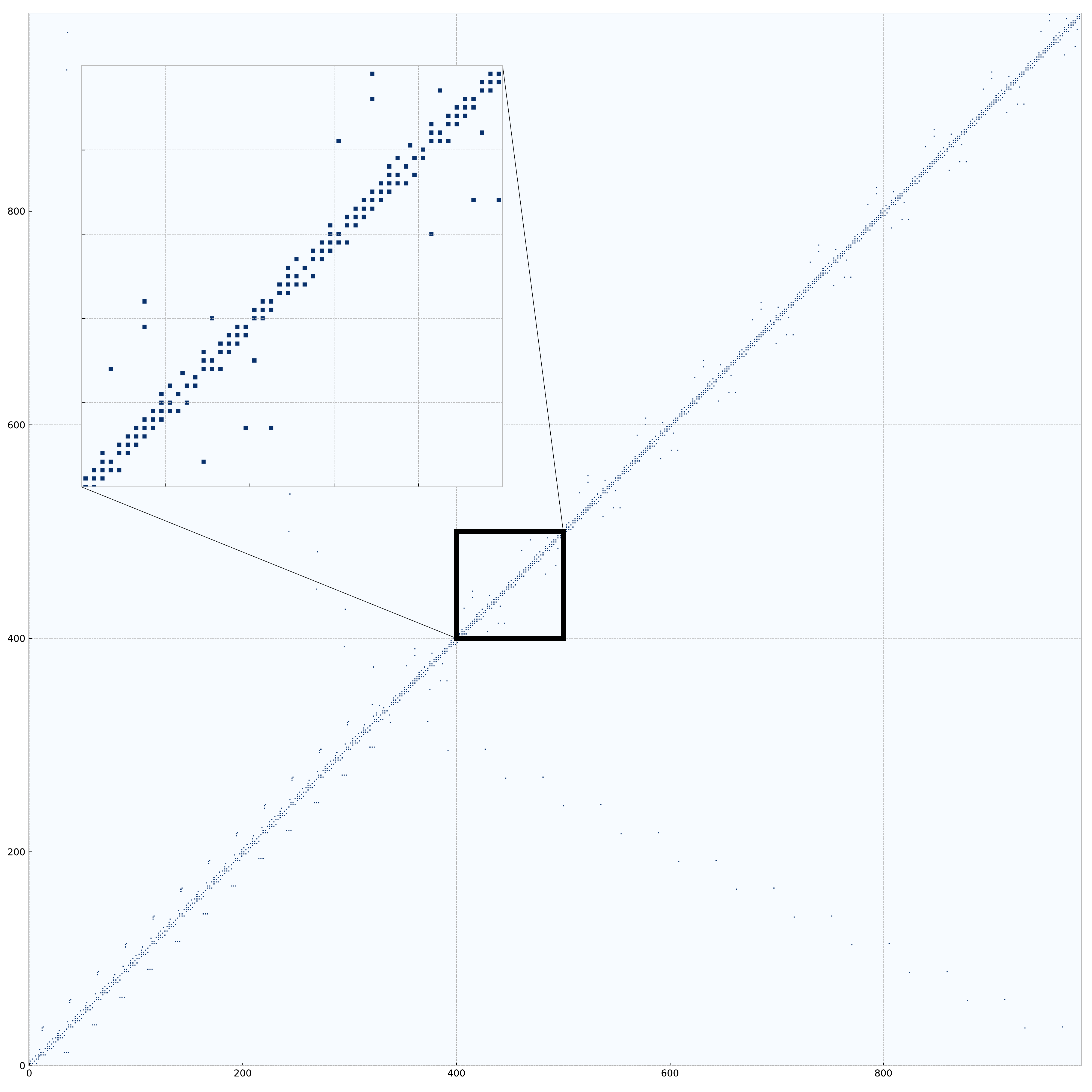}
     \end{subfigure}
     
     \begin{subfigure}[b]{0.15\textwidth}
         \centering
         \includegraphics[width=\textwidth]{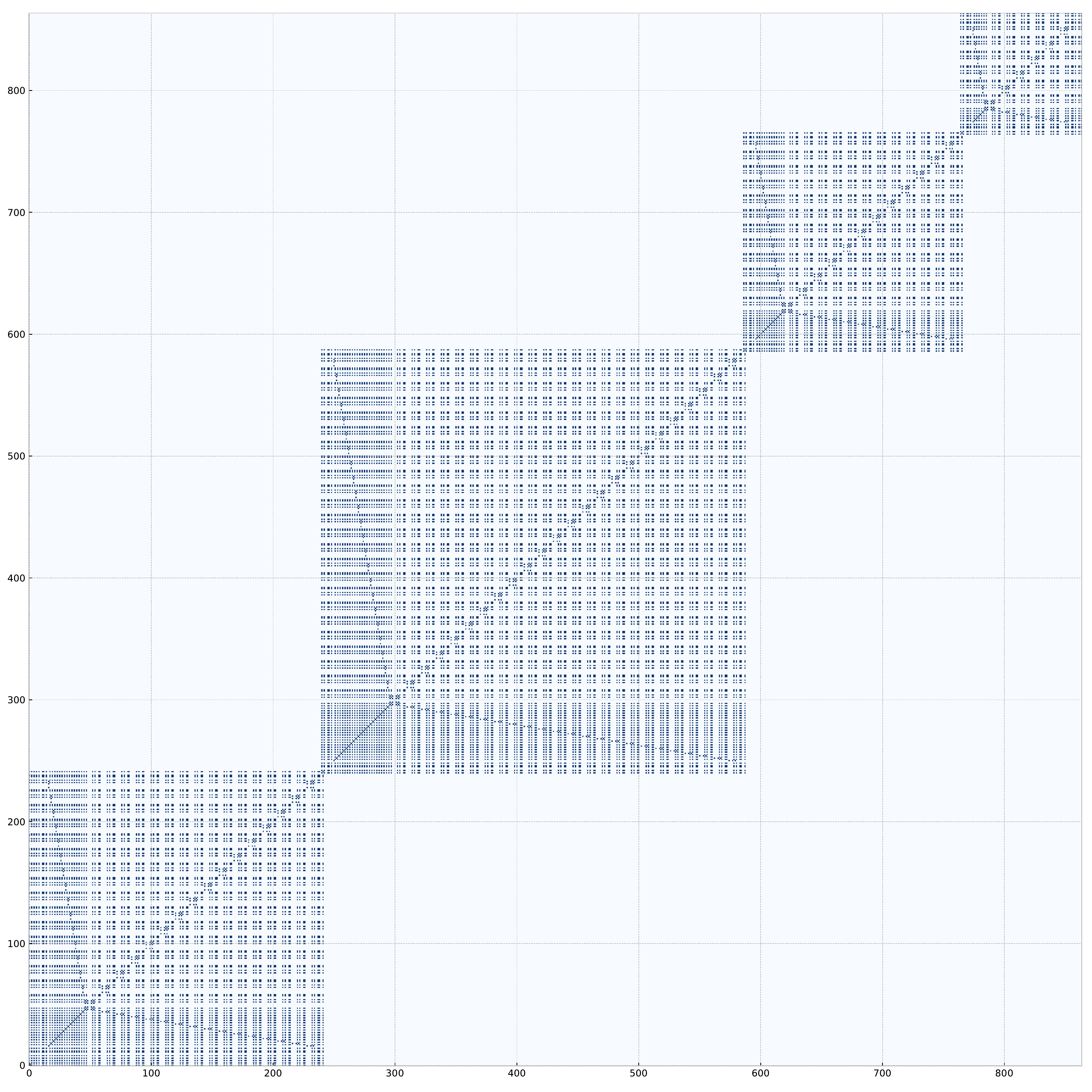}
     \end{subfigure}
     \begin{subfigure}[b]{0.15\textwidth}
         \centering
         \includegraphics[width=\textwidth]{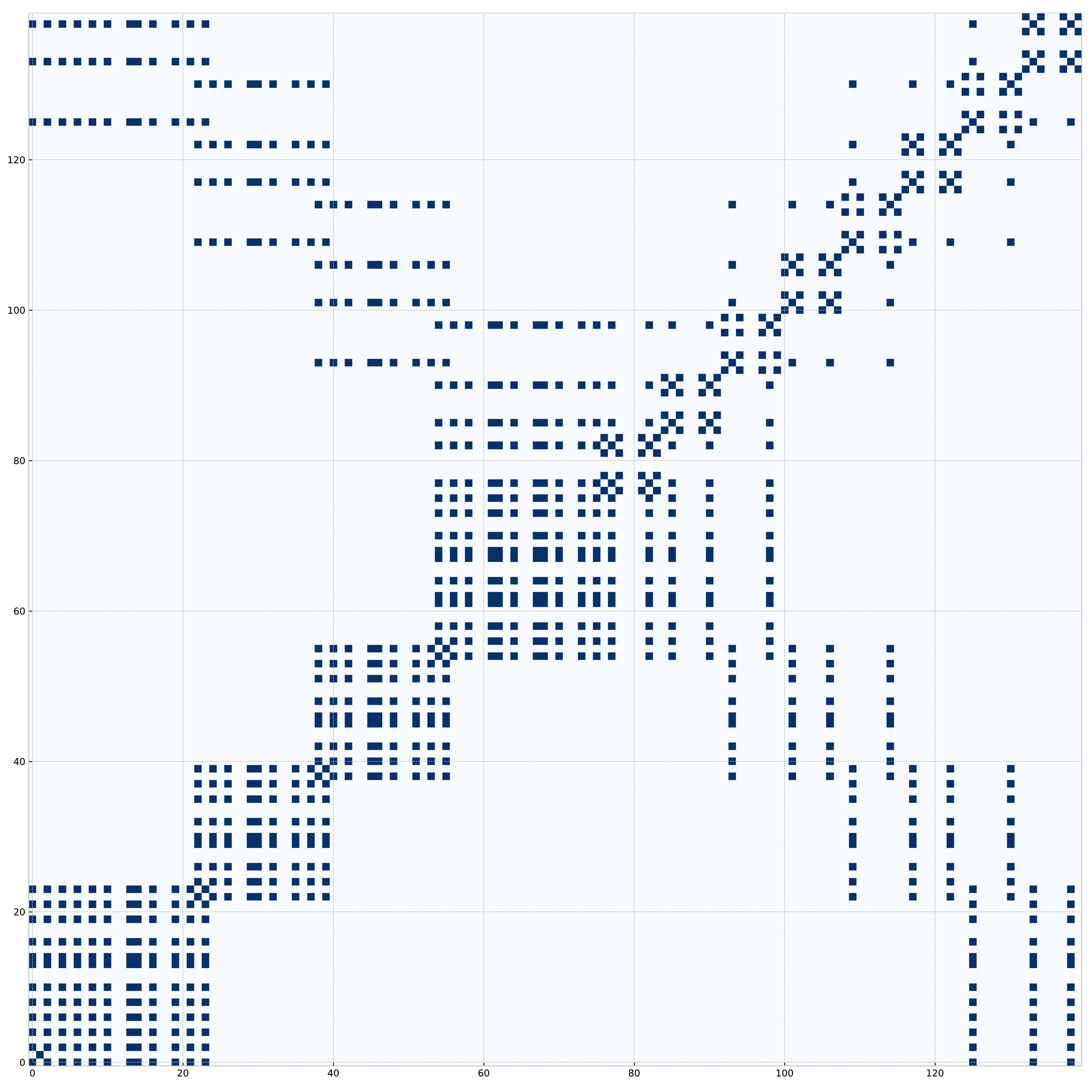}
     \end{subfigure}
     \begin{subfigure}[b]{0.15\textwidth}
         \centering
         \includegraphics[width=\textwidth]{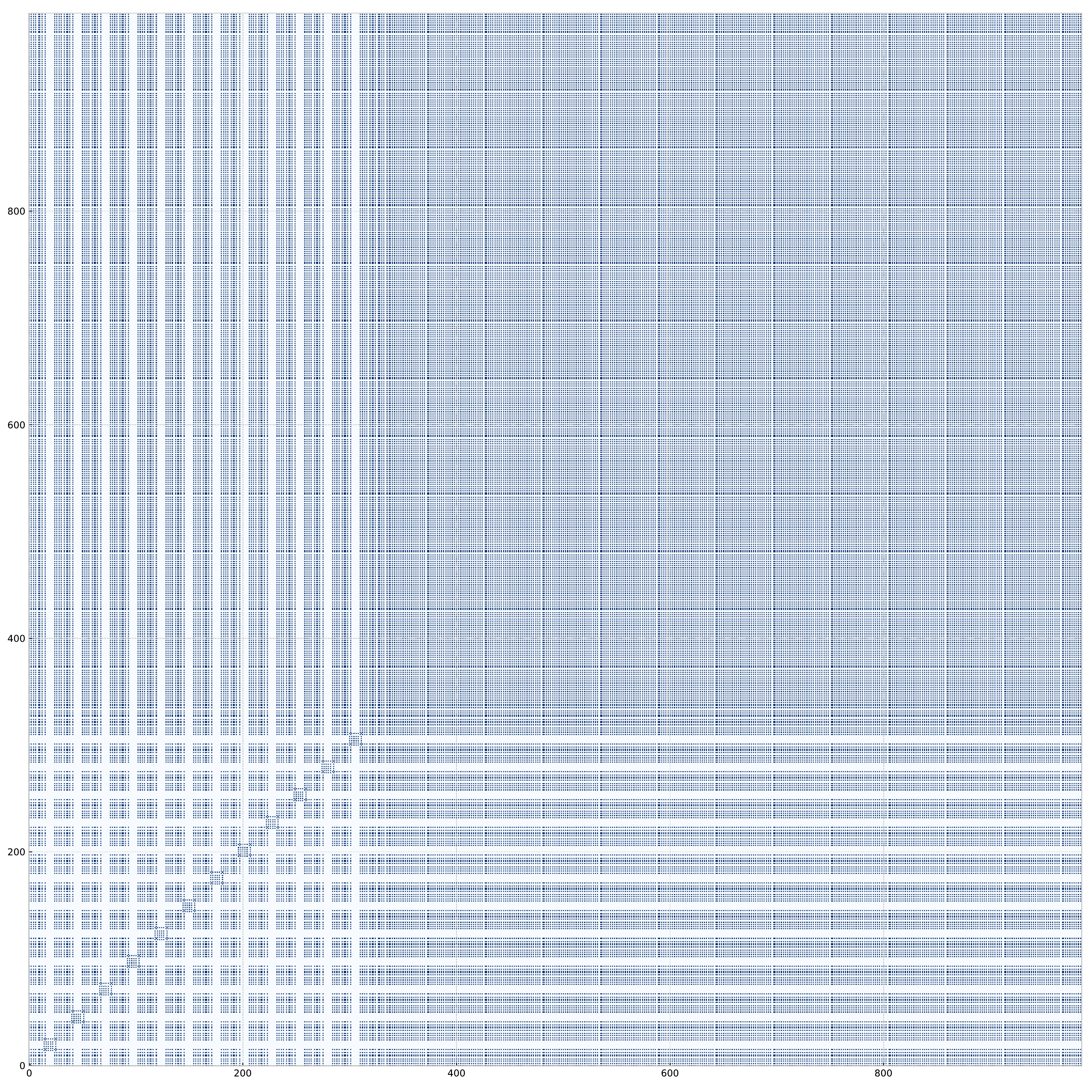}
         
     \end{subfigure}
    \caption{Dependency graphs (top) and the derived grouping schemes (bottom) for DenseNet-121, ResNet-18 and ViT-Base.}
    \vspace{-2mm}
    \label{fig:grouping_viz}
\end{figure}

\paragraph{Visualization of DepGraph.} 
Pruning large neural networks presents a considerable challenge due to the intricate process of grouping parameters. However, by employing the DepGraph, all coupled groups can be effortlessly obtained. 
We provide visualizations of the DepGraph $D$ and the derived grouping matrices $G$ for DenseNet-121~\cite{huang2017densely}, ResNet-18, and Vision Transformers~\cite{dosovitskiy2020image} in Figure \ref{fig:grouping_viz}. The grouping matrices are derived from the DepGraph as outlined in Algorithm \ref{alg:grouping}, where $G[i, j]=1$ signifies that the $i$-th layer belongs to the same group as the $j$-th layer. DenseNet-121 demonstrates a strong correlation between layers within the same dense block, leading to large groups during the structural pruning. The proposed Dependency Graph proves to be helpful when dealing with complex networks, as manually analyzing all dependencies in such networks is indeed an intractable task.

\paragraph{ImageNet.} 
Table \ref{tbl:imagenet_pruning} presents pruning results on ImageNet for several architectures, including ResNet, DenseNet, MobileNet, ResNeXt, and Vision Transformers. The target of this work is not to provide state-of-the-art results for various models, thus we only use the most basic importance criterion in this work. We show that a simple norm-based criterion, when combined with dependency modeling, can achieve comparable performance to modern approaches that use powerful criteria~\cite{liu2021group,you2019gate} and training techniques~\cite{wang2020neural}.

\begin{table}[t]
  \centering
  \resizebox{\linewidth}{!}{
  \small
  \begin{tabular}{c l | c c | c c}
      \toprule
      \bf Arch. \& Data & \bf Method & \bf Base & \bf Pruned  & \bf $\Delta$ &  \bf Speedup \\
    \hline  
      \multirow{4}{*}{\shortstack{LSTM\\(AGNews)}} 
      & DepGraph+Random & 92.10 & 91.23 & -0.87 & 16.28$\times$   \\
      & DepGraph+CP~\cite{li2016pruning} & 92.10 & 91.50 & -0.60 & 16.28$\times$ \\
      & Ours w/o SL & 92.10 & 91.53 & -0.57 & 16.28$\times$ \\
      & Ours & 92.10 & \bf 91.75 & \bf -0.35 & 16.28$\times$ \\
      \midrule
      \multirow{5}{*}{\shortstack{DGCNN\\(ModelNet40)}} 
      & DepGraph+Random & 92.10 & 91.05 & -1.05 & 10.05$\times$   \\
      & DepGraph+CP~\cite{li2016pruning} & 92.10 & 91.00 & -1.10 & 10.05$\times$ \\
      & DepGraph+Slim~\cite{liu2017learning} & 92.10 & 91.74 & -0.36 & 10.35$\times$ \\
      & Ours w/o SL               & 92.10 & 91.86 & -0.24 & 11.46$\times$ \\
      & Ours                 & 92.10 & \bf 92.02 & \bf -0.08 & 11.98$\times$ \\
      \midrule
      \multirow{4}{*}{\shortstack{GAT\\(PPI)}} 
      & DepGraph+Random & 0.986 & 0.951 & -0.035 & 8.05$\times$   \\
      & DepGraph+CP~\cite{li2016pruning} & 0.986 & 0.957 & -0.029 & 8.05$\times$ \\
      & Ours w/o SL & 0.986 & 0.953 & -0.033 & 8.26$\times$ \\
      & Ours   & 0.986 & \bf 0.961 & \bf -0.025 & 8.43$\times$ \\
      \bottomrule
  \end{tabular} 
  }
  \caption{Pruning neural networks for non-image data, including AGNews (text), ModelNet (3D Point Cloud) and PPI (Graph). We report the classification accuracy (\%) of pruned model for AGNews and ModelNet and micro-F1 score for PPI.}
  \vspace{-2mm}
  \label{tbl:other_datasets}
\end{table} 

\paragraph{Text, 3D Point Cloud, Graph and More.} In addition to CNNs and Transformers, our method is easily applicable to other architectures as well.  This part consists of experiments on a variety of data, including texts, graphs, and 3D point clouds, as shown in Table \ref{tbl:other_datasets}. We utilize LSTM for text classification by studying the effectiveness of DepGraph on recursive structures in which parameterized layers are coupled due to the element-wise operations. DepGraph is also tested on Dynamic Graph CNNs that contain aggregation operations for 3D point clouds. Furthermore, we conduct experiments with graph data, which require entirely different architectures from those used for other tasks. In this experiment, we concentrate on the acceleration of Graph Attention Networks, which have several coupled layers within each GNN layer. Considering the lack of works concerning pruning on these datasets, we combine DepGraph with some classic pruning methods in CNNs to establish our baselines. The results indicate that our method can be indeed generalized to a wide variety of architectures.

\section{Conclusion}
In this work, we introduce Dependency Graph to enable \emph{any structural pruning} on a wide variety of neural networks. Our work is the first attempt, to our knowledge, to develop a general algorithm that can be applied to architectures, including CNNs, RNNs, GNNs, and Transformers. 

\section*{Acknowledgment} 
This research is supported by the National Research Foundation Singapore under its AI Singapore Programme (Award Number: AISG2-RP-2021-023).

{\small
\bibliographystyle{ieee_fullname}
\bibliography{egbib}
}

\end{document}